\documentclass[runningheads]{llncs}
\usepackage{graphicx}
\usepackage{times}
\usepackage{bbm}
\usepackage{epsfig}
\usepackage{amsmath}
\usepackage{amssymb}

\begin{document}
\title{Cross Your Body: A Cognitive Assessment System for Children}
\author{Saif Sayed \and
Vassilis Athitsos}
\institute{Vision-Learning-Mining Lab, University of Texas at Arlington, Arlington TX 76013, USA 
\email{saififtekar.sayed@mavs.uta.edu, athitsos@uta.edu}
}
\maketitle              
\begin{abstract}
While many action recognition techniques have great success on public benchmarks, such performance is not necessarily replicated in real-world scenarios, where the data comes from specific application requirements. The specific real-world application that we are focusing on in this paper is cognitive assessment in children using cognitively demanding physical tasks. We created a system called Cross-Your-Body and recorded data, which is unique in several aspects, including the fact that the tasks have been designed by psychologists, the subjects are children, and the videos capture real-world usage, as they record children performing tasks during real-world assessment by psychologists. Other distinguishing features of our system is that it's scores can directly be translated to measure executive functioning which is one of the key factor to distinguish onset of ADHD in adolescent kids. Due to imprecise execution of actions performed by children, and the presence of fine-grained motion patterns, we systematically investigate and evaluate relevant methods on the recorded data. It is our goal that this system will be useful in advancing research in cognitive assessment of kids. 
\keywords{Action Recognition  \and Cognitive Assesment \and Action Segmentation}
\end{abstract}

\section{INTRODUCTION}

Mental illness can cause several undesirable effects on a person’s emotional, mental or behavioral states\cite{world2001world}. It is estimated that around 450 million people are currently affected by mental health issues, including schizophrenia, depression, attention-deficit hyperactivity disorder(ADHD) and autism spectrum disorder (ASD)\cite{world2001world}. More specifically, ADHD, which is a psychiatric neurodevelopmental disorder found in children and young adolescents, can have its traces evident as early as age 6 \cite{cormier2008attention}. Such traces may include deficits in executive functions\cite{barkley1997behavioral} inhibiting them to perform mental processes like planning, organizing, problem-solving as well as managing their impulses, including working memory, cognitive flexibility and inhibitory control \cite{best2010developmental}. These developmental shortcomings causes detrimental effects not only in their school performances but also at a higher level, trigger many negative effects in family, employment and community settings which can result into several socio-economic problems, causing low self-esteem and self-acceptance \cite{dendy2008executive}

While current methods use fMRI or sMRI scans\cite{riaz2018deep}, facial expressions\cite{jaiswal2017automatic} or clinical notes\cite{dai2018assessing},these methods provide good prognosis of the subject's cognitive condition at the brain activity/blood flow level, but are expensive and not portable. Embodied cognition tackles this problem with an understanding that our sensorimotor experiences with our social and physical environment helps in developing and shaping our higher cognitive processes\cite{wellsby2014developing}. Inspired by this approach, research \cite{mcclelland2014predictors} adopted the Head Toe Knee Shoulder(HTKS) task to assess these psychometric measures of self-regulation through physical performance for 208 subjects using obtrusive wearable sensors. 

Similarly a recent work \cite{bell2021activate} created a system called ATEC whose scores showed significant correlation with concurrent measurements of executive functions and significant discriminant validity between At-risk children and Normal Range children on multiple pre-existing tests like the CBCL Competency, CBCL ADHD Combined score, BRIEF-2 Global Executive Composite, BRIEF-2 Cognitive Regulation Index and SNAP-IV ADHD Combined Score. They measure psychometric scores such as Response Inhibition, Self-Regulation, Rhythm and Coordination which constitutes the Executive Function(EF) Score and Working Memory Index Score. These scores provide valuable information for differentiating adolescent kids susceptible to ADHD as compared to normal\cite{krieger2018assessment}. They used human annotators to evaluate the activities performed by kids, while current systems like \cite{sayed2019cognitive}, \cite{gattupalli2017cognilearn} use computer vision technique to detect these activities, but they do not produce scores that can be translated to produce these psychometric scores. The main contribution of the paper is to create a system that can produce an automated score of rhythm and accuracy, which is the fundamental component of creating the psychometric measures by utilizing the recording and scoring protocol followed ATEC system for the cross your body task and compare it with the human scores. The data has been recorded in real-time in an indoor environment, and shows children performing fine-grained activities. In this system, the children follow instructions to touch, using each hand, a specific body part (ear, shoulder, knee, or hip) on the other side of the body. 

\section{RELATED WORK}

Neuroimages like fMRI or sMRI have been traditionally used as clinical data for applications such as identification of ADHD \cite{zou2017deep,riaz2018deep} which use CNN to identify local spatial patterns of modulations of blood flow in a section of brain. While there are compelling research methods that can separate kids with ADHD from control subjects, these techniques require costly acquisition of brain scans and face the issue of portability. Instead of learning such information at a micro-level, one can study the effects of the disorder at macro-level, as human movements and how they can be affected by hyperactivity and/or inattention\cite{american2013diagnostic}. This inspired several wearable sensor based approaches\cite{hotham2018upper,kam2011high} and a significant difference was evident between non-ADHD controls and ADHD patients, but such methods require obtrusive sensors.

With the advent of sophisticated activity recognition systems, such human movements can be tracked and was first tried on adults using HTKS task\cite{gattupalli2017cognilearn}. The prior work most related to ours is the method presented in \cite{sayed2019cognitive}, which was also applied to videos of children performing the Cross-Your-Body task. 

Our work has significant differences, and advantages, compared to \cite{sayed2019cognitive} and can be used for fully-automated scoring of the children's performance. The method of \cite{sayed2019cognitive} cannot be used on its own for fully automated scoring, due to two limitations which our method overcomes: the first limitation was that it required human annotations to convert a child's performance into several segmented videos, each segment corresponding to a specific instance of the hand moving to touch a body part. Second, the output of the system in \cite{sayed2019cognitive} simply classified each video segment, without providing frame-level labels indicating when exactly the subject touches the instructed body part. Frame-level labels are necessary for scoring the rhythm and timing of each child's performance, which is an important aspect of the human experts' evaluation protocol.

\section{Data Acquisition and Protocol Definition}

The goal of the system is to facilitate the development of an automated scoring environment, whose output correlates as much as possible with scores produced by human experts for the same videos. Simulated datasets have been previously created for similar tasks using adults\cite{gattupalli2017cognilearn}, but they lack the unique motion dynamics that the children participants display in our recording.

\subsubsection{Data Recording} 
The subjects were recorded at multiple indoor locations and strict quality control was maintained (such as keeping the distance from the subjects to the camera within a specified range), to ensure consistent acquisition quality and every kid was given the same instruction. The data was recorded using Kinect V2. A screen is used to display a music video where the host instructs the kids to perform the task following the song of "Cross-your-body". There are 5 tasks overall and each task has varying sub-segment. In a single sub-segment, the subject is told to perform 3 touches based on the announced body part using the hand from the opposite side and alternate sides, for example "Cross your body touch your hips, hips, hips. For first 2 tasks the subject is required to touch the same body part as instructed, but for the other 3 tasks the challenge becomes cognitively demanding as they are told to follow opposite rules, for example in task 3 subject requires to touch ears when instructed to touch knees and vice versa. Similarly task4 switches shoulders and hips and task5 includes rules of both task 3 and 4. The scoring system is followed according to the research done by Bell \cite{bell2021activate}

\subsubsection{Scoring Scheme}

 The objective of our data processing module is to apply activity segmentation algorithms to evaluate the performance of the participants. The scoring protocol created by the psychologist experts specifies 2 scores: accuracy and rhythm. The accuracy score depends on the amount of times that the subject touches the desired body part correctly, and the rhythm score depends on the amount of times that the subject touches the desired body part within one second after receiving the instruction. These two scores signify different psychometric measures necessary for measuring self-regulation, response inhibition, working memory, co-ordination and attention \cite{bell2021activate}. Thus, our system is designed to produce both those scores.

For the accuracy score part, the goal is to detect if the relevant activity is happening or not in a video. A potential approach is to manually segment and annotate these videos, to ensure only one activity happening per video, and then to recognize the activity in each trimmed video. It has limited real-world use, as it requires significant manual effort to produce the trimmed video. Furthermore, as this approach produces a video-level class label, as opposed to frame-by-frame labeling, it could not be used for computing rhythm score, which requires identifying the time when  the hand touches the respective body part.

To address these limitations, in this paper we follow a problem formulation where the system output should predict both the body part that is touched, and the time during which it is touched by the hand. The input to the system is a video segment, which typically has more than one action. For the training examples, the ground truth includes frame-level labels. The input video segment is provided automatically by the video capturing system, with no need for manual annotations, based on the time that each instruction is provided to the child. These instructions are pre-recorded, and the time that they are issued is known to the system.

Based on the above considerations, we formulate our problem as an frame-level supervised action segmentation problem, similar to formulations applied on the MPIICooking2 dataset\cite{fayyaz2020sct} to understand fine-grained activities. We note that our problem could also possibly be tackled as an activity localization problem, but in our work so far we have not pursued that approach. 

\subsubsection{Annotation Scheme:} 

\begin{figure}[t]
\begin{center}
  \includegraphics[width=0.65\linewidth]{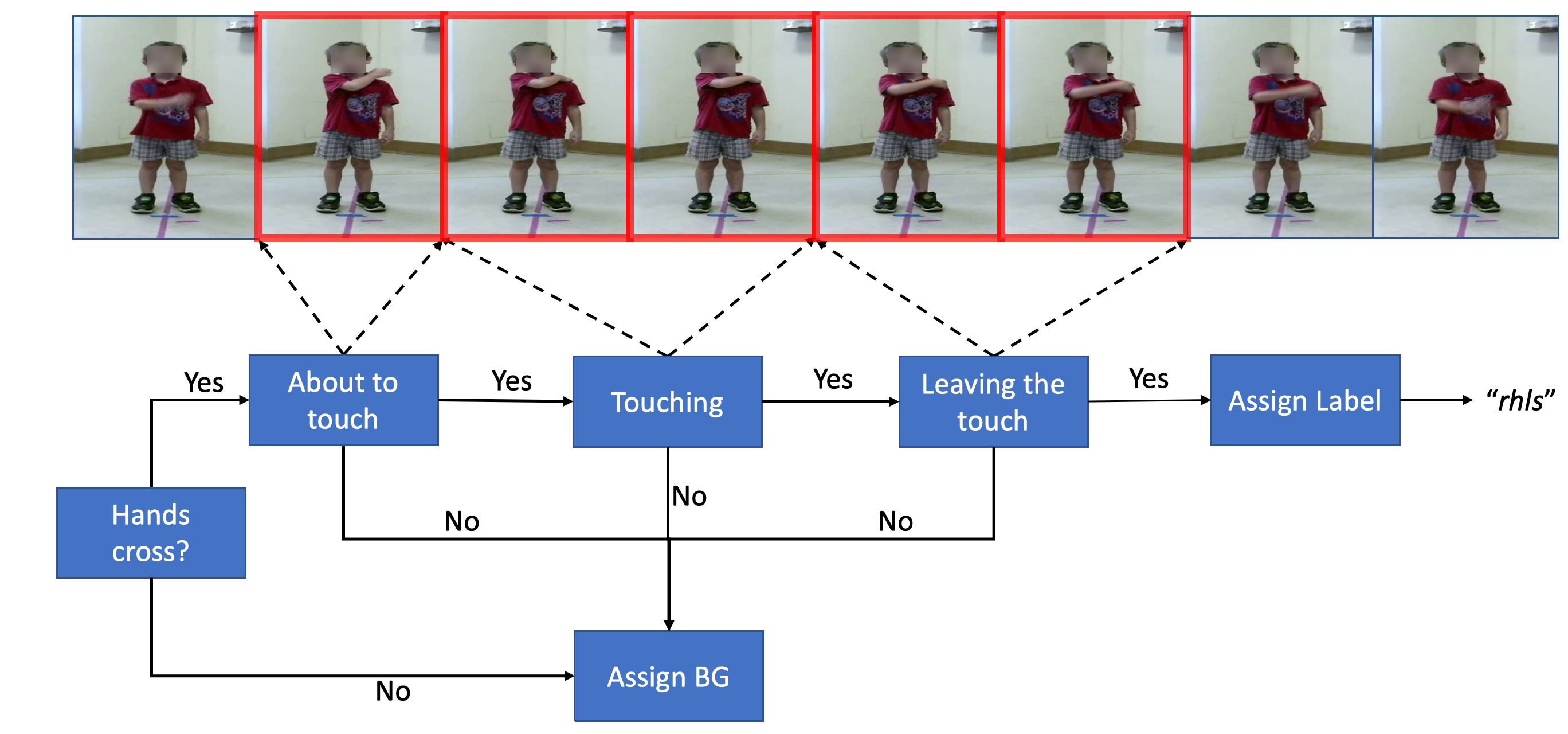}
  \caption{Annotation Scheme for the Cross Your Body System. Highlighted Red frames were given annotation as right hand left shoulder(rhls)}
\end{center}
\label{fig:annot_scheme}
\end{figure}

The annotation scheme we adopted in our system is illustrated in Fig. \ref{fig:annot_scheme}. Here we  annotate explicitly those frames where the hand approaches the body part, then touches, and finally leaves the body part. As seen in the figure, a gesture is considered valid only if the hand has crossed the midline of the body and fulfilled the 3 sequential steps. The steps and the corresponding frames are highlighted in the figure. We first identify the time segment when the hand is about to touch the body part. We chose an approximate distance of a few inches around the designated body part. In the subsequent time segment the subject touches the body part, and in the third time segment the hand leaves the body part and gets to a distance of few inches from it. A video segment of video is assigned the appropriate label if it fulfills all these steps, else it is designated as background(BG). A strict protocol was followed as there are cases where a subject keeps the hand crossed and near the body part while using the other hand to touch the other side of the body part.

\subsubsection{Dataset Statistics} Overall the dataset consists of 894 total videos recorded for 19 subjects, and has on an average 2.7 activities per video. The average length of a video sample is around 3.3 seconds, while the maximum and minimum length of the videos are around 3.6 and 3.1 seconds respectively. There are around 2500 activities in these videos having durations ranging from 0.03 seconds to 1.36 seconds. The dataset consists of 8 classes without background. The class-wise distribution of the dataset is as shown in the table \ref{data_stats}. There are 8 classes indicating the combination of hand and body part. The 4 lettered class label is comprised of the first and second letter indicating the side: left(l) or right(l), the second letter stands for hand(h) and the fourth letter stand for the body part: ear(e), shoulder(s), hip(h), knee(k).

\begin{table}[h]
\begin{center}
\footnotesize\setlength{\tabcolsep}{2.3pt}
\caption {Class-wise duration distribution of the dataset.\label{data_stats} }
\begin{tabular}{ c|c|c c c|c}
Class label & Index & Min(sec) & Max(sec) & Mean(sec) & Sample Count\\\hline	
lhre & 0 & 0.03 & 0.93 & 0.21 & 319 \\\hline			
rhle & 1 & 0.03 & 0.90 & 0.23 & 224 \\\hline					
lhrs & 2 & 0.03 & 1.17 & 0.20 & 388 \\\hline						
rhls & 3 & 0.03 & 0.87 & 0.20 & 285 \\\hline							
lhrh & 4 & 0.03 & 1.37 & 0.25 & 338 \\\hline						
rhlh & 5 & 0.03 & 1.00 & 0.24 & 257 \\\hline						
lhrk & 6 & 0.03 & 0.83 & 0.19 & 337 \\\hline						
rhlk & 7 & 0.03 & 0.87 & 0.21 & 284 \\\hline	
\end{tabular}
\end{center}
\end{table}

\begin{figure*}
    \includegraphics[width=0.99\linewidth]{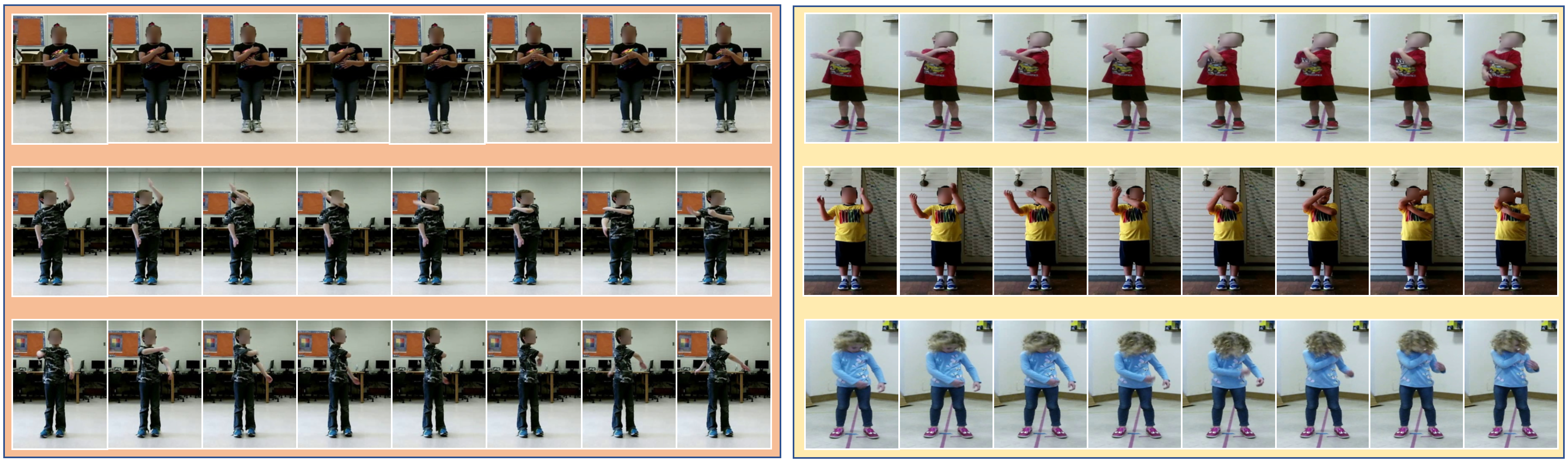}
    \caption{Examples of action instances in Cross Your Body Data. The left part shows instances belonging to categories within the set of touching-shoulder, from top to bottom "right hand left shoulder", "left hand right shoulder", "right hand left shoulder", illustrating the level of variations that the touching-shoulder action can have. On the right the first 2 samples from top to bottom illustrate examples of touching-shoulder and touching-ear instances in which one hand occludes another, while the third sample illustrates occlusion by head while touching the shoulder.}
    \label{fig:data_overview}
\end{figure*}

\subsubsection{Data Properties} The recorded data has several attractive properties that distinguish it from the existing datasets.
\textit{High Quality:} All the videos were recorded in Full HD resolution and were recorded in indoor conditions that ensured good illumination quality.

\textit{Richness and Diversity:} The only practice that the subjects received was showing them once a video illustrating how to touch each body part. There were no instructions given to the children during the recording phase to improve their gestures or motion patterns. This resulted in a very realistic dataset that has many unique and novel motion patterns, as can be seen in figure \ref{fig:data_overview}. The dataset has high intra-class variation of and also has occlusions during performance of these activities. Also the diversity in the speed at which a subject touches a body part varies drastically, resulting into cases where the touch of a body part spans just a few frames.

\textit{Fine-grained action differences:} Since there are classes like touch ear v/s touch shoulder, this dataset is unique from the point of view of inter-class variations, as there are samples belonging to different classes where the body pose looks very similar. Furthermore, since the resolution of the hands is relatively small, hand detection and tracking is a challenge, thus posing a unique use-case for activity detection and recognition algorithms.

\section{System Definition} 
The goal of the system is to recognize and score the fine-grained actions in this dataset. Given the scoring requirements, the system is essentially an action segmentation system. Note that this dataset can also be used to evaluate activity localization algorithms, but we have so far not pursued that approach.  Action segmentation involves frame-level predictions of an untrimmed video that may contain one or more activities. 

 We systematically evaluated 3 action segmentation methods based on Temporal  Convolution  Networks(TCN), namely MSTCN++\cite{li2020ms}, ASRF\cite{ishikawa2021alleviating} and DTGRM\cite{wang2020temporal} on the data and also included one pose based activity recognition system ST-GCN\cite{yan2018spatial} to understand the significance of pose. Training protocols follow the original paper unless stated otherwise. 

For action segmentation, the setup involves a training set of N videos, where each video is composed of frame-wise feature representations $x_{1:T} = (x_1,...,x_T)$, where $T$ is the length of the video. Using these features the system outputs the predicted action class likelihoods $y_{1:T} = (y_1,...,y_T)$, where $y_t\in R^C$ and $C$ is the number of classes. During test time, given only a video, the goal is to predict $y_{1:T}$.

All of the experiments were performed using a user-independent 6-fold cross validation system 
where it was ensured that, for each split, there is no training video of any person appearing in the test set. 
The remaining section explains the feature extraction and the analysis on the performance of these methods.

\subsection{Feature Extraction} The actions in the video are highly dependent not only on the motion patterns but also on the appearance information. Due to the fine-grained nature of the activities like touching ear v/s touching shoulder, missing on the latter information will lead to incorrect prediction. We used the 2 modalities of data, mainly optical flow and RGB frames to produce intermediate frame-representation using I3D network\cite{carreira2017quo} pretrained on Kinetics dataset. We have chosen a temporal window of 16 frames to compute the I3D features. The I3D features extracted from each modality is concatenated together to produce a feature representation $\mathbf{x}_i\in \mathbb{R}^{2048\times T_i}$, where $T_i$ is the length of the video $i$

\subsection{Action Segmentation}
We chose Temporal Convolution Networks(TCN)-based modelling systems, because TCNs have a large receptive field and work on multiple temporal scales, and thus they are capable of capturing long-range dependencies between the video frames.   
The reason we chose this multi-scale option is because the instruction given usually follows a theme. For example, if the instruction is "Cross your body touch your Ears, Ears, Ears", the subject is expected to touch ears three times (each time with the opposite hand than the previous time). Consequently, frame-level predictions become easier if the network understands that the actions happening in the video are related to ears, and that hands alternate.  

MSTCN++ is an improvement over MSTCN where the system generates frame level pedictions using a dual dilated layer that combines small and large receptive field in contrast to MSTCN\cite{farha2019ms}. While MSTCN++ has the ability to look at multiple temporal fields, it still lacks the ability for efficient temporal reasoning. This drawback was resolved in DTGRM where they used Graph Convolution Networks(GCN) and model temporal relations in videos. While models like MSTCN++ and MSTCN use smoothing loss to avoid over-segmentation errors, this method introduced an auxilliary self-supervised task to encourage the model to find correct and in-correct temporal relations in videos.DTGRM and MSTCN++ both work predict frames directly and there is no concept of detecting action boundaries. Since in our dataset and task it is necessary to understand when an action starts and ends, and also since the motion is so fine-grained, it is important to accurately detect action boundaries. To resolve this problem, we also employed another method, ASRF, which alleviates over-segmentation errors by detecting action boundaries. For analysis of the importance of the method based on the modality of the data, we trained each method on 3 different modalities: I3D features extracted on RGB frames, I3D features extracted on flow frames, and a third modality consisting of concatenating the I3D features of the first two modalities.

Evaluation metrics: For evaluation, the metrics we employ are framewise accuracy(Acc), framewise accuracy without background(Acc-bg), segmental edit distance (Edit), and segmental F1 score measured at overlapping thresholds of 10\text{\%}, 25\text{\%} and 50\text{\%} denoted by F1-10, F1-25 and F1-50 respectively. The overlapping threshold is based on the metric of Intersection over Union (IoU) ratio. We added the framewise accuracy without background as a metric since a major portion of the frames in the dataset is background.

We did 2 forms of analysis: event-based analysis, where models were trained using all 9 classes including background. The second analysis was done based on a subset of the labels. More specifically we relabeled "left hand right ear" and "right hand left ear" to just ear and similarly for shoulder. This resulted in only 3 classes: ear, shoulder and background.

Table \ref{allclass_results} illustrates the performance of all the 3 segmentation models for all the classes. It can be observed that ASRF shows the best performance as compared to MSTCN++ and DTGRM, as it is able to predict action boundaries. Note that the metric of \emph{Acc-BG} is more important than \emph{Acc} as the dataset has a lot of background frames. More importantly, the flow modality produced better results as compared to RGB or even concatenated I3D features, and this is further discussed in the analysis subsection. Other fine-grained activity datasets like \cite{shao2020finegym} have shown similar observations where they illustrate RGB values contribute less due to subtle differences between classes as compared to coarse-grained classes.

We have also investigated how data modality plays a role on capturing temporal dynamics in actions that are very fine. For this, we chose a subset of classes like touching ears and shoulders because they are spatially separated by a very low margin as compared to touching hip and knee. Table \ref{subsetclass_results} illustrates the performance of all the 3 segmentation models for this subset of classes. It is evident that flow plays an important role in understand these motion patterns.

\begin{table}[h]
\begin{center}
\footnotesize\setlength{\tabcolsep}{2.3pt}
\caption {Performance of Action Segmentation models for All Classes.\label{allclass_results} }
\begin{tabular}{ c | c | c c | c | c c c}
Method	& Modality	& Acc & Acc-BG & Edit & F1-10 & F1-25 & F1-50\\    \hline
\scriptsize{MSTCN++}        & RGB & 79.59 & 28.43 & 67.94 & 69.53 & 67.13 & 55.07 \\\cline{2-8}
                                      & Flow & 82.73 & 36.10 & 71.15 & 74.13 & 71.81 & 60.13 \\\cline{2-8}
                                      & Both & 82.11 & 36.24 & 71.68 & 74.71 & 72.15 & 59.90 \\\hline
\scriptsize{DTGRM}          & RGB & 81.10 & 26.75 & 64.83 & 68.59 & 66.09 & 54.75 \\\cline{2-8}
                                      & Flow & 83.55 & 35.76 & 71.91 & 75.35 & 72.52 & 61.05 \\\cline{2-8}
                                      & Both & \textbf{83.67} & 35.68 & 70.50 & 75.07 & 72.50 & 60.74 \\\hline
\scriptsize{ASRF}           & RGB & 63.34 & 36.42 & 55.73 & 59.90 & 54.64 & 42.79 \\\cline{2-8}
                                      & Flow & 80.45 & \textbf{48.66} & \textbf{73.94} & \textbf{78.04} & \textbf{75.56} & \textbf{63.66} \\\cline{2-8}
                                      & Both & 80.98 & 44.03 & 73.45 & 77.10 & 74.81 & 63.27\\\hline
\end{tabular}
\end{center}
\end{table}

\begin{table}[h]
\begin{center}
\footnotesize\setlength{\tabcolsep}{2.3pt}
\caption {Performance of Action Segmentation models.(Set Level: Ear and Shoulder)\label{subsetclass_results} }
\begin{tabular}{ c | c | c c | c | c c c}
Method	& Modality	& Acc & Acc-BG & Edit & F1-10 & F1-25 & F1-50\\    \hline
\scriptsize{MSTCN++}        & RGB & 82.41 & 33.95 & 71.92 & 73.51 & 70.38 & 59.34 \\\cline{2-8}
                                      & Flow & 86.00 & 51.36 & \textbf{80.44} & 82.45 & 80.03 & 69.76 \\\cline{2-8}
                                      & Both & 85.87 & 48.08 & 78.87 & 81.76 & 79.90 & 69.00 \\\hline
\scriptsize{DTGRM}          & RGB & 84.00 & 34.22 & 72.39 & 75.65 & 73.00 & 61.79 \\\cline{2-8}
                                      & Flow & \textbf{86.69} & 50.80 & 79.09 & 82.62 & 80.85 & 70.23 \\\cline{2-8}
                                      & Both & 86.09 & 43.45 & 74.76 & 80.08 & 77.43 & 66.66 \\\hline
\scriptsize{ASRF}           & RGB & 83.28 & 44.22 & 75.23 & 77.34 & 75.74 & 65.88 \\\cline{2-8}
                                      & Flow & 85.02 & \textbf{56.55} & 79.95 & \textbf{82.90} & \textbf{80.86} & \textbf{70.50} \\\cline{2-8}
                                      & Both & 84.63 & 49.12 & 77.79 & 80.64 & 78.93 & 68.69 \\\hline
\end{tabular}
\end{center}
\end{table}

\begin{figure}[t]
\begin{center}
  \includegraphics[width=0.65\linewidth]{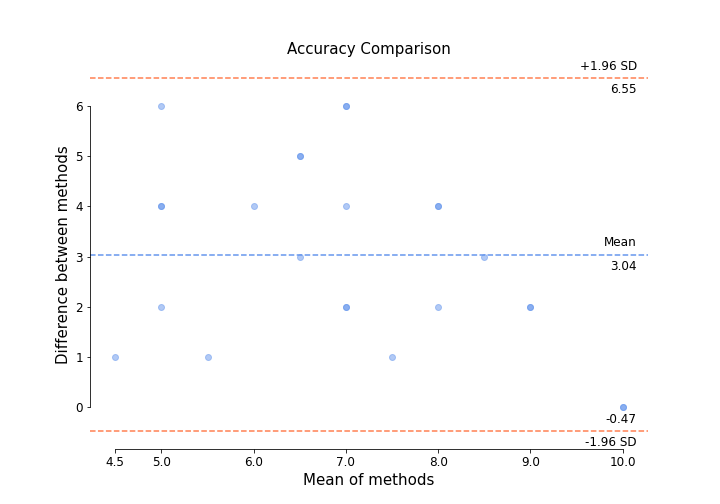}
\end{center}
  \caption{Accuracy Score comparison of Human and machine scores}
\label{fig:acc_compare}
\end{figure} 

\subsection{Analysis}
\subsubsection{Comparison with human scores} 
\begin{figure}[t]
\begin{center}
  \includegraphics[width=0.65\linewidth]{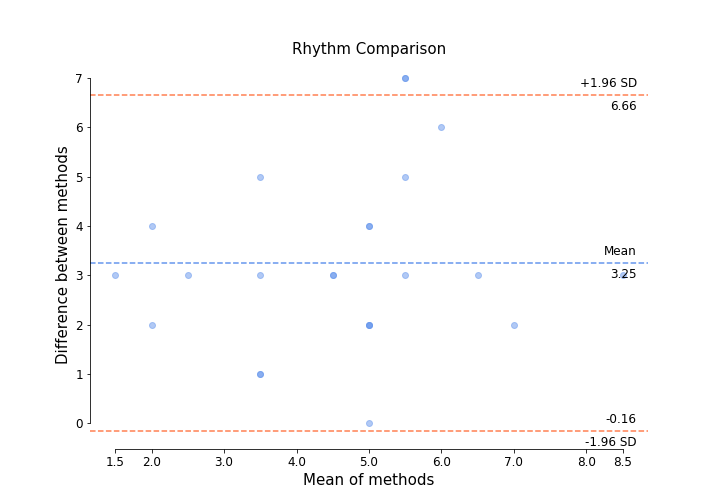}
\end{center}
  \caption{Rhythm Score Comparison of Human and machine scores}
\label{fig:rht_compare}
\end{figure}

Using the segmentation results and the times when the instruction was made by accessing the video, the relevant accuracy and rhythm scores for each subject was produced by the system for all the 5 tasks and all subjects. The Bland-Altman plots(Fig. \ref{fig:acc_compare} and \ref{fig:rht_compare}) shows the comparison of the system scores and human scores for the activities performed by the kids. The Y-axis indicates the difference of the scores generated by machines and humans and the X-axis indicates the mean of the scores using both methods. Every point in the scatter plot indicates the measurement for a user which may overlap. For each plot, The blue line indicates the mean of difference of human and machine scores (estimated bias) which is 3.04 for accuracy metric and 3.25 for rhythm. The red lines refers to interval (mean ± 1.96 × standard deviation) which signifies the limits of agreement between human and machine scores. For accuracy the limit of agreement is [6.55,0.47] and [6.66,0.16] for rhythm. Ideally the mean of differences should be closer to 0. This shows that there is a potential of improvement in reliably detecting the touches. 

\begin{figure}[t]
\begin{center}
  \includegraphics[width=0.65\linewidth]{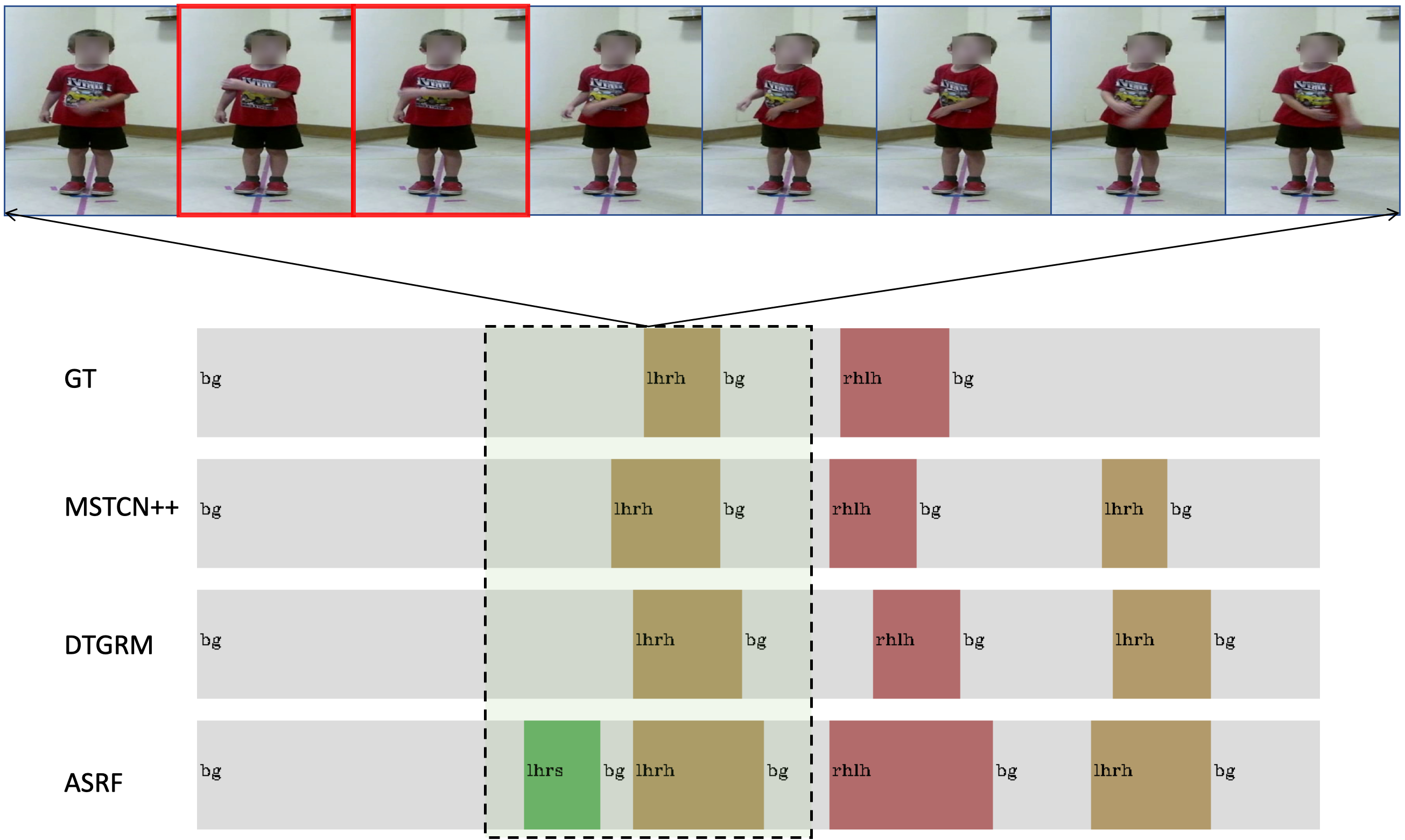}
\end{center}
  \caption{Segmentation Results for different methods. ASRF incorrectly predicts lhrs(highlighted section) when the subject hovered his hand around shoulder and then decided to touch right hip}
\label{fig:analysis_2}
\end{figure} 

\subsubsection{What cannot be handled by the current models}
To illustrate the limitations of the current methods, we provide the segmentation results 
of some cases where the network failed. 

\textit{1) Self correction:} The network cannot always handle the scenario where the user self-corrects the touch. For example in Fig.\ref{fig:analysis_2}, the subject first intends to touch his shoulder. Then, in the middle of moving the hand towards the shoulder, he corrects himself and proceeds to touch the hip. Such cases are essential in this dataset and the target application, as the subject has to utilize their working memory to decide which body part to touch based on the instruction and type of the rule they are told to follow. The task was intentionally designed by the psychologist experts so that subjects can get easily confused and need to self-correct. The segmentation results show that the best system ASRF incorrectly predicted \textit{lhrs} as it failed to understand that the touch did not happen. 

\textit{2) Confusion between ear and shoulder:} While the pose system clearly illustrated that the system cannot handle spatially fine-grained poses like touch ear v/s touch shoulder for some cases, this issue was echoed in action segmentation results as well, which used much more sophisticated I3D features. 

\textit{3) Intense motion:} Sometimes out of confusion and haste to complete the task, the subject performs touching of body parts at a high speed, and that makes it challenging to predict. 

\textit{4) Occlusions:} As the subjects perform action very quickly, this results into scenarios where the activities overlap and hand from the previous action occludes the other hand which is being used to perform the next action, causing the network difficulty to track.

\subsection{Conclusion}
In this paper, we introduce a system for Cross-Your-Body task focusing on cognitive assessment using kids as subjects. The system differs from existing works as there is a direct comparison between the scores provided by human experts and machines. The recorded data provides diverse activities which have high intra-class variability and low inter-class variability. It also includes many unique and realistic actions that involve uncoordinated motion patterns that vary in pace and has occlusions. We have empirically investigated significance of pose and I3D features and different data modalities by viewing it as an action segmentation problem. Many interesting findings show that the current state of the art systems find it difficult to recognize these activities. Our system demonstrates creation of 2 fundamental metrics required to measure several executive functions and shows promising potential for future research. This system can be used as a non-intrusive solution for cognitive assessment in kids where there is no need of an expert to manually score the cognitively demanding tasks. 

\subsubsection*{Acknowledgements}
This work was partially supported by National Science Foundation grants IIS 1565328.

{\small
\bibliographystyle{splncs04}
\bibliography{main}
}

\end{document}